# Can Social Ontological Knowledge Representations be Measured Using Machine Learning?

Ahmed Izzidien[1]

[1] The Psychometrics Centre, the University of Cambridge.

## Abstract

Personal Social Ontology (PSO), it is proposed, is how an individual perceives the ontological properties of terms. For example, an absolute fatalist would arguably use terms that remove any form of agency from a person. Such fatalism has the impact of ontologically defining acts such as 'winning', 'victory' and 'success' in a manner that is contrary to how a non-fatalist would ontologically define them. While both the said fatalist and non-fatalist would agree on the dictionary definition of these terms, they would differ on specifically how they can be brought about. This difference between the two individuals can be induced from their usage of these terms, i.e., the co-occurrence of these terms with other terms. As such a quantification of this such co-occurrence offers an avenue to characterise the social ontological views of the speaker. In this paper we ask, what specific term co-occurrence should be measured in order to obtain a valid and reliable psychometric measure of a person's social ontology? We consider the social psychology and social neuroscience literature to arrive at a list of social concepts that can be considered principal features of personal social ontology, and then propose an NLP pipeline to capture the articulation of these terms in language.

## Introduction

### Measuring Personal Social Ontology

In forming sentences, words are typically selected for their subjective suitability in transferring meaning. However, it is accepted that meaning is never based on individual words, but the association each word has with the next. These associations are grounded in a knowledge of the subjective appropriateness of making one or more relations between each word's ontological features. For example, comparing the phrase 'time will deal a blow to the unjust' to the phrase 'one ought to standup firm in the face of injustice', the former uses the inanimate, agentless, uncontrollable object of 'time', in relation to the imposition of justice. Whereas the latter uses agency, immediacy, power and control in relation to it. The former can be said to represent a form of disempowerment through fatalism, whereas the latter represents a form of empowerment through free will. The rationales of the sentences may be initially stated as, justice will establish itself *vs.* justice requires human intentional agency to be established.

Each of these can be said to represent a rationale carried by their respective authors. In these two cases, the rationale informs the reader of *how* the text's author perceives concepts of justice and agency. This perception, or knowledge, is manifested in their chosen word associations. Associations that are made by the author's mind between the *subjective* ontological features of the words. For justice, despite its shared definition for both authors, does not functionally operate, or interface, with other elements in one author's perceived reality in the same way that it does for the other author's.

A further approach is the consideration of how concepts in sentences are associated with each other. Thus, in a sentence on deterring plagiarism, for example, is the follow-up sentence one that appeals to a threat of power, or integrity? These associations can possibly inform us of an author's own subjective assessment of the appropriateness of concepts linkages. In doing so, they may further inform of a unique rationale.

However, it is necessary to identify the principal dimensions of personal social ontology in order to know which terms one should measure. Such a measure would require the qualification of orthogonal dimensions that represent such knowledge.

We consider social psychology literature to discover principle dimensions of human behaviour such as the propensity to be social and cooperate (Dovidio et al., 2017; Milinski et al., 2002; Moghaddam, 2000) generated in part through processes such as theory of mind (Bogaert et al., 2008; Dawes et al., 2012; Hu et al., 2016), empathy (Lamm et al., 2011; Patil & Silani, 2014) and a social comparison heuristic (Civai, 2013; CorradiDell'Acqua et al., 2013; Decety & Cowell, 2017).

From a social neuroscience perspective, we detail findings related to social perception:

**An Ontogenetic Account**

Theory of mind is seen to facilitate not only giving behaviour but a comparison heuristic. Both implicated in the prefrontal cortex (PFC). Additionally, results have been seen to provide evidence for the existence of two distinct systems: one reflecting self-serving assessment supported by the ACC and mPFC and a second that appears to identify (distributive) injustice regardless of the target by invoking the anterior insula (aINS). This concept of self-regarding and other regarding has been theorised by (Bogaert et al., 2008) as central in differentiating the choice to give or withhold in a decision. The fact that the insula is directly involved in physiological, food, and pain-related processing supports the general notion that prosocial behaviour, is implemented on a fundamental physiological level similar to breathing, heartbeat, hunger and pain (Dawes et al., 2012).

In terms of activation, a meta-analysis of 107 neuroimaging studies of self- and other-related judgments using multilevel kernel density analysis assessed processes used to perceive and understand the self and others. Relative to non-mentalising judgments, both self and other judgments were associated with activity in the mPFC, ranging from ventral to dorsal extents, as well as common activation of the left TPJ and posterior cingulate (Denny et al., 2012). Activation of the TPJ being linked with orienting attention to salient information, encoding agency, and applying temporary states onto others (Eres et al., 2018).

This TPJ activity occurs consistently when reasoning about the thoughts, intentions, and beliefs of others as well as when differentiating the self from others. Such thus suggests that the role of the TPJ in moral judgment and decision making is to discern whether a behavioural outcome is appropriate based on the agent's mental state. This view is also supported by previous research linking TPJ activity with retrospective justifications of behaviours and mentalising capabilities (Kliemann et al., 2008), (Eres et al., 2018), (Van Overwalle, 2009).

A meta-study also determined that a core-network including mPFC and bilateral posterior TPJ was activated across a range of distinct tasks, supporting the idea of a core network for theory of mind that is activated whenever we are reasoning about mental states, irrespective of the task- and stimulusformats. Considering an Activation Likelihood Estimation (ALE) meta-analysis of neuroimaging studies, which assessed 2,607 peak coordinates from 247 experiments in 1,790 participants, it was found to determine that the brain areas that were consistently involved in moral decisions showed more convergence with the ALE analysis targeting theory of mind versus empathy (Bzdok et al., 2012).

Findings on theory of mind as a facilitator of giving within studies involving children have been considered throughout the literature. It has been argued that children who possess advanced theory of mind are more likely to act prosocially, yet the empirical findings are mixed (Hao & Liu, 2016), (Lucas et al., 2008). In an attempt to address this issue definitively, a meta-analytic integration of 76 studies including 6,432 children between 2 and 12 years of age, collapsed across all studies, discovered a significant association ($r = 0.19$) which indicated that children of higher theory of mind scores also had higher scores on concurrent measures of prosocial behaviour. The magnitude of this effect was similar across theory of mind assessments requiring the identification of others' cognitions versus emotions, and it existed irrespective of whether the theory of mind measure imposed demands on false belief reasoning or not (Imuta et al., 2016). The opposite manipulation has also indicated that theory of mind is a relevant factor in these scenarios. Oxytocin produces enhancements of prosocial relevant perception by increasing theory of mind related neural activations (Hu et al., 2016).

It has been found that the ACC, whether indexed by hemodynamic activity or by frontal scalp negativities, shows an enhanced response to unfair offers, specifically when making decisions for oneself (Alexopoulos et al., 2012), (Gabay et al., 2014). In contrast, the aINS appears to respond equally to unfairness regardless of perspective (Civai et al., 2012), (Corradi-Dell'Acqua et al., 2013). This aINS response need not necessarily mirror a purely emotional reaction to unfairness (Tabibnia et al., 2008), (Sanfey et al., 2003). Instead, rejecting inequitable distributions appears to reflect a cognitive heuristic, a psychological anchor, that can be easily adjusted when salient contextual cues enter the environment (Decety & Yoder, 2017), (Civai, 2013).

In this respect, PCC activation has been considered as reflective of the situational adequacy (Robertson et al., 2007), (Pujol et al., 2008), (Pujol et al., 2008), i.e. context. In a study which examined this, strategic fair offers in the UG elicited enhanced activity in prefrontal areas and altruistic fair offers in the DG evoked increased activity in the ACC and the PCC. The latter indicated that the proposers do not give up the social norm of fairness even in the condition of non-punishment (Chen et al., 2019) a finding also demonstrated using ERPs, whereby proposers were shown not to give up the fairness norm, despite their highly unfair proposals (Chen et al., 2019).

Both the aforementioned ACC and insula are thought to play a key role in the affective component of empathy as demonstrated in a meta-analytic analysis (Lamm et al., 2011).

Empathy is also known to facilitate an inherent aversion to harm through processes creating social coherence through the sharing of affective states in order to be motivated to help, and not harm others (Roberts et al., 2019). A review demonstrates that a contextual appraisal, including perceived fairness may modulate empathic neuronal activations (Bernhardt & Singer, 2012). A meta-analytic study also determined the AI and ACC as both being implicated in the experience of pain and the witnessing of pain (Cui et al., 2015) as also demonstrated in the meta-analysis by (Lamm et al., 2011).

Using a multilevel kernel density analysis, a whole brain based quantitative meta-analysis of recent fMRI studies of empathy was performed. This analysis identified the dACC-aMCC-SMA and bilateral anterior insula as being consistently activated in empathy. Yet, alone, empathy incentivization has been determined to not always lead to more pro-social allocations. Although an empathy manipulation has previously successfully increased dictators' feelings of empathy towards recipients, the dictators' decisions on how to split the money were not affected (Lönnqvist & Walkowitz, 2019). Thus, towards our theory, an interdependence of factors of theory of mind, empathy, a fairness heuristic and context, are driving the decision process. A final note is to be made on the prevalent research framework on neural activities, whereby inferences are made based on the manipulation of perception. Implying the brain works as a receiver of data, which it then *reacts to*. An alternative frame is presented by (Buzsaki, 2019, p. 57,59.) which considers the brain as already generating data, pre-perception.

Whereby internal projection, estimation and reaction are primary, while perceived information takes a more secondary route.

Given the salience of *social survival* being present both in pro-socials and pro-selfs, it will be the centre of our focus.

**Measure development**

No current method exists to measure personal social ontology. Some foundational work has been carried out in the field of social ontology (Porpora, 2009; Searle, 2006, 2008), although its focus has been more on the nature of social entities as represented by language than on the individual's own perception of the nature of social entity interfaces as represented through their unique use of language.

In terms of measuring PSO using digital methods, a number of challenges must still be met. One being that textual associations are often causal and consequential in nature. These have been challenging to discern using current natural language programming (NLP) and machine learning methods (Ittoo & Bouma, 2011; Li & Mao, 2019; Yang & Mao, 2014). Thus, an automated measure would require an advancement of existing methods. Yet a number of current approaches have been successful in measuring other personal constructs (Boyd & Pennebaker, 2016) with greater precision and speed than that of self-measure survey questions (Boyd et al., 2015). These have included automated measures of personality (Arnoux et al., 2017; Kosinski et al., 2013; Park et al., 2015), social values (Boyd et al., 2015; Wilson et al., 2016), thinking dimensions (Chung & Pennebaker, 2008) and mental profile mapping (Boyd, 2018).

Research from hundreds of labs have repeatedly found that various categories of language are direct reflections of personality-relevant psychological processes (Tausczik & Pennebaker, 2010), (Hai-Jew, 2017), and that aspects of a person's mental life can be adequately modelled from even modest language samples (Pennebaker & King, 1999) (Arnoux et al., 2017), (Azucar et al., 2018), (Boyd, 2018). Boyd and Pennebaker have developed a process known as Mental Profile Mapping (MPM), in which they identify a broad consistency of a person's traits (Boyd & Pennebaker, 2017), (Boyd, 2018). They have further developed a Meaning Extraction Method (MEM) to identify psychologically meaningful themes by recognising words that repeatedly co-occur across a corpus. The process uses statistical methods through which they undertake a principal component factor analysis of self-description essays. The themes that emerge are then treated as independent dimensions of thought ("Revealing Dimensions of Thinking in Open-Ended Self-Descriptions," 2008). The MEM is also seen to share many features with Latent Semantic Analysis which examines similarities between texts using singular value decomposition on the occurrence of key content words (Yu et al., 2008), (Tonta & Darvish, 2010). Other methods employing the same for textual summarisation (Gaikwad & Mahender, 2016).

It may be argued, that using the principal factors that represent social ontology to feature engineer a text, so that the ML algorithm is able to capture the most pertinent dimensions, may be an efficient method at mitigating unnecessary processing of spurious and irrelevant factors. The question remains, what are the principal factors that best explain personal social ontology?

Acts of cooperation are central to human behaviour (Trivers, 1971), (Milinski et al., 2002). The formation of secure social bonds has been considered a fundamental human need (Tabibnia & Lieberman, 2007). Based on the social psychology literature, and drawing on earlier work (Izzidien, 2021; Izzidien & Stillwell, 2021) it may be argued that humans have two principal social senses: *survival* and that of *cooperation*. In sum, the ability of humans, to recognise harm aversion in others, has an effect of generating a meta-quality towards them. This meta-quality has the effect of *deterring* the observer from harming the observed person, and concomitantly affords the observed person a degree of *protection* from being harmed. These bias the observer to act pro-socially. Consequentially these feelings have the same qualities of a sense of *responsibility* not to harm and concomitantly a *right* not to be harmed. While these senses do not generate these as explicit social values, the senses consequently generate the essential qualities of these said values.

Yet, as with all biases, they do not necessarily determine outcome. Twin gene research has determined that genes contribute to approximately half of the variance in measures of self–report altruism, empathy, nurturance and aggression, including acts of violence. Furthermore, using a 22–item social responsibility questionnaire with 174 pairs of monozygotic twins and 148 pairs of dizygotic twins, the results demonstrated that 42% of the reliable variance was due to the twins' genes, 23% to the twins' common environment and the remainder to the twins' non–shared environment (Rushton, 2004). A secondary consideration is therefore necessary, that is, the degree to which the evaluator chooses to act in accordance with or contrary to their bias. That is, their social volition. This bias can be considered an outcome of the interaction between the two innate senses of survival and cooperation. That is, a bias that nudges humans towards sociality. The sense of *ought* is inferable when considering the human reference. The human reference being the instinct to be a social species, and to avoid socially costly, damaging decisions. Whereby a cost is perceived in acting in a manner which is antithetical to them.

It may be suggested that here consistency in acting in accordance with such in-built instincts for social survival breeds consistency, i.e., cooperation begets survival, which begets cooperation, and avoids the costly dissonance of acting in a manner that is contrary.

Such bias, when present in humans, serves to symmetrically facilitate socialisation. Whereby individuals possess a sense of *ought*, towards each other.

The perspective of each agent to the other, each mentalistically aware of each other's bias, arguably creates a situation where an expectancy to act in a manner that each would find proper presents itself. This is because, the bias to be fair is symmetrical, not only projected towards one person. It generates an expectancy in the other person. As such any perception of a legitimate need, by wither agent, is construable as a right. However, the social value of a claim-right is not explicit, its essential qualities are emotionally present as a result of the said bias.

If we were to represent such a perception abstractly, one may summarise it in the following: Humans typically use a sense (e.g., perception), a degree of power (e.g., agency), with an intention (e.g., purpose) to interact. The outcome can be qualified as one that the perpetrator would wish to receive themselves or not. A common term that describes this at a basic level is *fairness*. Whereby an act is perceived as fair if one would be satisfied were it to be done to themselves (Izzidien & Stillwell, 2021). The principle abstract components of this value of fairness, are a fulfilment of the rights one is due, and concomitantly, the responsibilities. A person is then able to implicitly evaluate if these have been fulfilled, and thus implicitly evaluate the consequence as positive rewards or negative retributions.

To summarise these features, once would consider a text for its representation of:

> Power, purpose, the rights, and duties due, as well as the outcome and its consequence of the power interaction in terms of rewards or punishments.

It may be argued that such perceptions can form the starting point for the features. To present such a mapping, the paper proposes that a connectome to map these factors would allow for a cognitively based digitized representation. This is in contrast to other representational techniques that use network analysis to map all topics in a discourse, without recourse to, or the integration of, a theory on social language construction. Machine learning on graphs with manually annotated data representations as baselines would allow a machine learning algorithm to then differentiate the different categories of each personal ontology.

**Considerations of the methodologies premise**

The premise used in the paper is that mentalisations effect language, however another theory advanced by Whorf (1956) is that language effects how we understand, or mentalise, the world around us. This assumption is also made by some critical discourse analysis researchers, who hold that crosslinguistic differences alter perceptual and conceptual event representations. Such has recently been revived in several neo-Whorfian theories of how perception and cognition are connected to language (Boroditsky, 2001). Proponents of these theories argue that categorical distinctions made in one's language result in language-specific categorical distortions in objective perceptual tasks (Margolis & Laurence, 2015, p. 328), (Burdick, Mihalcea, Boyd, & Pennebaker, 2019). Such crosslinguistic differences, if accepted, may indeed impinge on the final representation, yet, given that social survival is a human instinct, and that the pro-social bias theory accounts for cross-cultural influences (Izzidien, 2021; Izzidien & Stillwell, 2021), it may be argued that the principal *pro-social* features will materialise irrespective. It also stands that the bias, given its innateness, will be present, to some degree, despite varied socialisation.

**Further work**

As a use case for a PSO psychometric measure, an accurate qualification and quantification of this form of knowledge representation, and a capture of its stability through time, could allow for its incorporation as a necessary, and as yet largely absent dimension in artificial intelligence (AI) reasoning methodologies. Especially as it represents a form of reasoning that all humans necessarily undertake, be it deliberately or otherwise.

It may be noted that current Large Language Models (LLMs), in being trained on data written by multiple authors (Rong, 2014) can conflate the different ontological associations, made by the personal perceptions and social positions of each one of the authors, leading to a new, and possibly epistemologically incoherent one, one that is certainly *artificial*.

Further work on testing the validity and reliability of the suggested PSO dimensions is necessary using traditional psychometric methods. Once established, the aid principal factors will be usable to measure the said dimensions from an author's text.

The development of a PSO can serve several purposes, namely, a new dimension of measuring implicit biases, a characterisation of power perceptions and the ability to map principle human social features from textual data.

**Acknowledgments**

I would like to express my gratitude to the Psychometric Centre, the University of Cambridge, and the Social Decision-Making Laboratory, the University of Cambridge, for facilitating resources needed for undertaking parts of this study.